\documentclass{article}
\usepackage{PRIMEarxiv}

\usepackage[utf8]{inputenc} 
\usepackage[T1]{fontenc}    
\usepackage{hyperref}       
\usepackage{url}            
\usepackage{booktabs}       
\usepackage{amsfonts}       
\usepackage{nicefrac}       
\usepackage{microtype}      
\usepackage{lipsum}
\usepackage{fancyhdr}       
\usepackage{graphicx}       
\graphicspath{{media/}}     

\title{Performance Analysis of Machine Learning Algorithms in Chronic Kidney Disease Prediction
\thanks{\textit{\underline{Citation}}: 
\textbf{Iftekhar Ahmed, et al. (2022). Performance Analysis of Machine Learning Algorithms in Chronic Kidney Disease Prediction. Pages 1-6. DOI: 10.1109/ICICT54375.2022.9946591} 
}}

\author{
  Iftekhar Ahmed \\
  Department of CSE \\
  Leading University \\
  Sylhet, Bangladesh \\
  \texttt{iftekharif007@gmail.com} \\
  \And
  Tanzil Ebad Chowdhury \\
  Department of CSE \\
  Leading University \\
  Sylhet, Bangladesh \\
  \texttt{tanzilebad@gmail.com} \\
  \And
  Biggo Bushon Routh \\
  Department of CSE \\
  Leading University \\
  Sylhet, Bangladesh \\
  \texttt{routh.biggo123@gmail.com} \\
  \And
  Nafisa Tasmiya \\
  Department of CSE \\
  Leading University \\
  Sylhet, Bangladesh \\
  \texttt{nafisatasmiya388@gmail.com} \\
  \And
  Shadman Sakib \\
  Department of CSE \\
  Leading University \\
  Sylhet, Bangladesh \\
  \texttt{shadman@lus.ac.bd} \\
  \And
  Adil Ahmed Chowdhury \\
  Department of CSE \\
  Leading University \\
  Sylhet, Bangladesh \\
  \texttt{adil@lus.ac.bd} \\
}

\begin{document}
\maketitle

\begin{abstract}
Kidneys are the filter of the human body. About 10\% of the global population is thought to be affected by Chronic Kidney Disease (CKD), which causes kidney function to decline. To protect in danger patients from additional kidney damage, effective risk evaluation of CKD and appropriate CKD monitoring are crucial. Due to quick and precise detection capabilities, Machine Learning models can help practitioners accomplish this goal efficiently therefore, an enormous number of diagnosis systems and processes in the healthcare sector nowadays are relying on machine learning due to its disease prediction capability. In this study, we designed and suggested disease predictive computer-aided designs for the diagnosis of CKD. The dataset for CKD is attained from the repository of machine learning of  UCL, with a few missing values; those are filled in using "mean/mode" and “Random sampling method” strategies. After successfully achieving the missing data, eight ML techniques (Random Forest, SVM, Naïve Bayes, Logistic Regression, KNN, XG\_Boost, Decision Tree, and AdaBoost) were used to establish models, and the performance evaluation comparisons among the result accuracies are measured by the techniques to find the machine learning models with the highest accuracies. Among them, Random Forest as well as Logistic Regression showed an outstanding 99\% accuracy, followed by the Ada Boost, XG-Boost, Naive Bayes, Decision Tree, and SVM, whereas the KNN classifier model stands last with an accuracy of 73\%. 
\end{abstract}


\section{Introduction}
The kidney eliminates surplus fluid and waste from our 
bodies. The kidneys also eliminate acid created by our body's 
cells and maintain proper blood levels of salt, calcium, and 
other vital minerals. They are primarily responsible for 
removing toxins from the blood and turning waste into urine. 
Around 160 grams of weight and one to one and a half liters of 
urine are excreted daily by each kidney. Every 24 hours, the 
two kidneys work together to filter 200 liters of fluid out of the blood \cite{1}. It is a crucial component of our body. But 
occasionally, a sickness may attack it. If the kidneys do not 
remove waste that is impacted by toxins, kidney failure can 
result in mortality. The kidneys' role is to strain the blood by 
removing poisonous compounds from the body by utilizing the 
bladder through urination. Kidney problems can be categorized 
as either acute or chronic \cite{2}. In the global economy, kidney 
illness is a major issue. Two to three percent of the annual 
income of high-income countries is lost to this disease. 
Circumstances that affect the kidneys and lessen their ability to 
keep us healthy are included in chronic kidney illnesses. If a 
kidney condition worsens, waste can accumulate to the highest 
levels in our blood and lead to problems like hypertension, 
weak bones, anaemia, poor nutrition, and nerve damage. 
Additionally, kidney dysfunction increases the risk of cardiac 
and vascular diseases. 

CKD affects the human body globally and has a high rate 
of morbidity and mortality \cite{3}. Therefore, it is very important 
to predict and diagnose CKD in its early stages. It might make 
it possible for patients to receive timely care to slow the spread 
of the illness. Patients with CKD must also consider other 
factors, such as blood potassium, urea, calcium, phosphorus, 
and other elements when managing their diets. ML techniques 
are now frequently used in healthcare to create disease 
prediction models due to the availability of biomedical data. 
Additionally, approaches like ensemble and deep learning 
significantly increased the prediction power of machine 
learning models. Accurate illness prediction models can be 
created by extracting information from electronic health 
records (EHR) \cite{4} \cite{5}. Thus, with an adequate amount of data 
ML technologies can become a vital tool in predicting various 
renal diseases and can become a handful tool in providing 
medical attention to the patients who need that the most. 
However, from the standpoint of public health, laboratory data 
is still not widely accessible and that is hindering the 
development of computer-aided medical technology. 

A wide range of research is being conducted nowadays, from 
identifying and categorizing various diseases like cancer to 
heart, liver, and kidney disorders. To be more explicit, machine 
learning was regarded as the key technique in 2004 for 
identifying a variety of cancer-related issues \cite{6}. Thus, it is 
evident how drastically healthcare epidemiology is changing as 
ML applications for disease prediction become more 
widespread. Huang et al. \cite{7} studied two datasets to predict 
breast cancer and evaluated the effectiveness of single SVM 
classifiers as well as SVM classifier ensembles. Since the 
effectiveness of the SVM classifier can be significantly 
impacted by kernel functions, they combined a variety of 
different kernel functions in their research. The results showed 
that linear kernel-based SVM ensembles and RBF kernel-based 
SVM ensembles perform better on short and large datasets, 
respectively. Another remarkable work using the ML technique 
is done by Shah et al. \cite{8} to predict cardiovascular disease 
using several supervised algorithms. This study set out to 
identify a reliable predictor of cardiovascular disease among 
several machine learning techniques. By achieving 90.78\% 
accuracy on the training data, KNN outperformed the other 
algorithms in the proposed model. 

A prominent cause of death in recent years has been 
chronic kidney disease or CKD for short. Between 1990 to 
2017, there was an increase in CKD-related mortality around 
the world by 41.5\% \cite{9}. Therefore, diagnosing the illness or 
tracking down the cause can be a more sensible strategy for 
effective risk management. Random forest seemed to have the 
best performance among the six machine learning algorithms 
used by Qin et al. \cite{10} to evaluate kidney disease, with a 
diagnostic accuracy of 99.75\%. However, due to errors 
produced by the proposed model, the researchers developed a 
different model by combining logistic regression and random 
forest utilising perceptron, which demonstrated an accuracy of 
99.83\%. Chen et al. \cite{11} run multiple experiments to find out 
the best algorithms to identify CKD by utilising three separate 
modellings. During their initial experiment, KNN and SVM 
provided higher accuracy which is close to 100\% compared to 
SIMCA which performed slightly worse. In addition, their 
further experiment was done with composite data with added 
random noise on the same 3 algorithms and the result showed 
SVM provided 99\% accuracy. Charleonnan et al. \cite{12} designed 
a CKD identifier with the SVM algorithms in 2016. They 
experimented to determine the most accurate CKD 
identification, using many ML classifiers (SVM, KNN, logistic 
regression, and decision tree). A classification result of 98.3\%, 
98.1\%, 96.6\%, and 94.8\% were provided in each model, 
respectively. On the pre-processed datasets, Shamrat et al. \cite{13} 
analyzed the performance of supervised machine learning 
approaches such as Logistic Regression, Random Forest, Decision tree, and KNN algorithms to determine the best 
performing Kidney disease classifier. According to the 
research, Logistic Regression and Random Forest performed 
with a 100\% accuracy rate. A computer-aided disease predictor 
for kidney diagnosis was suggested by Imran et al. \cite{14}. On 
two different datasets, experiments were conducted utilizing 
three different machine learning (ML) techniques: wide and 
deep learning, feedforward neural networks, and logistic 
regression. The tests' findings showed that feedforward neural 
networks performed admirably on both balanced and 
imbalanced datasets.  

The aim of our work is to shed light on how CAD can aid 
in diagnosing chronic kidney disease and can help the 
healthcare system make better decisions. We also want to 
demonstrate that we used a number of supervised machine 
learning techniques and evaluated their effectiveness to find the 
best classifier that works with our model. The paper’s 
remaining portions are arranged as follows: A full approach 
including the dataset used, data preprocessing procedures, a 
brief explanation of the suggested methodology, and an 
explanation of various ML techniques are all included in 
Section II. Results and discussion as well as evaluation metrics, 
and findings comparison, are illustrated in Section III. Finally, 
After the concluding remark section IV closes.

\section{ Materials And Methods }
\begin{figure}[htbp]
    \centering
    \includegraphics[width=0.8\linewidth]{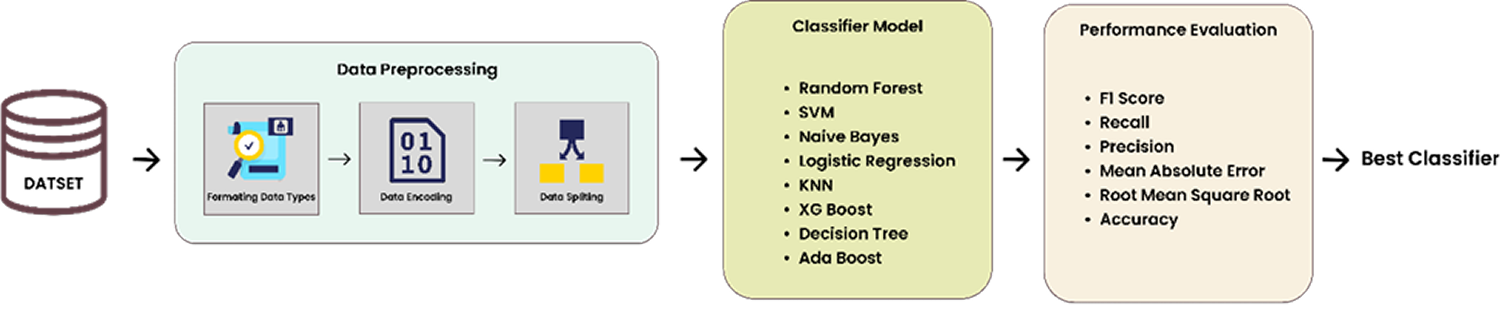} 
    \caption{Proposed Workflow for Chronic Kidney Disease  Prediction }
    \label{fig:methodology}
\end{figure}

This section is concerned with the research's theoretical 
soundness clarity. The study uses CKD data from UCI machine 
learning repository. Several preprocessing approaches were 
employed to clean this data. The features of the dataset were 
selected and after cleaning and processing, the data were 
separated into training and test sets. Training data was used to 
train eight machine learning classification algorithms. After 
training, the algorithms were used to predict test data. Fig.~\ref{fig:methodology}
portrays the overall workflow for evaluating the model in order 
to predict CKD.       

\subsection{Data Acquisition }

\begin{figure}[htbp]
    \centering
    \includegraphics[width=0.8\linewidth]{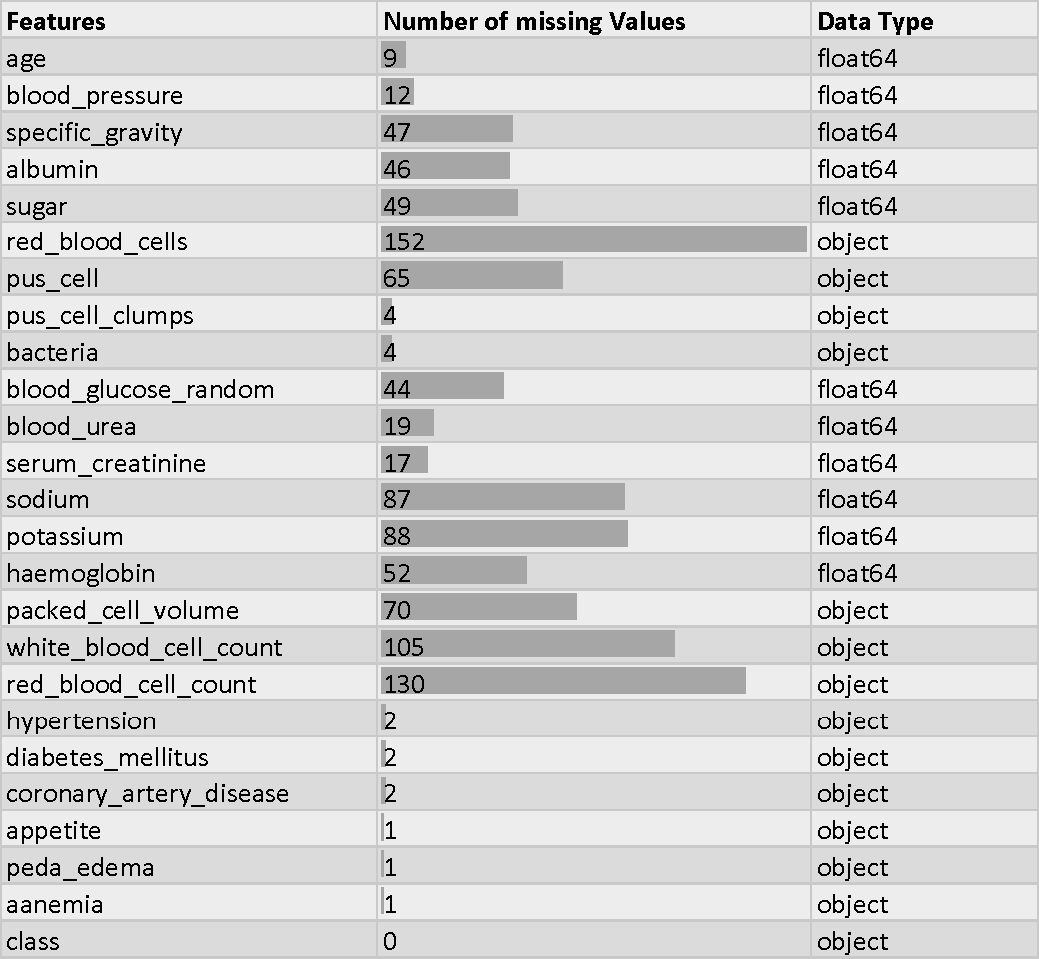} 
    \caption{ Overview of the chronic kidney disease (CKD) dataset }
    \label{fig: Overview of the chronic kidney disease (CKD) dataset }
\end{figure}

A well-known machine learning repository by the name of 
UCI is the source for our dataset \cite{15}. This dataset contains a 
total of 400 individual pieces of data. This dataset has a total of 
25 different features, where 250 are ‘CKD’ and 150 are ‘not 
CKD’. This dataset contains two types of data, i.e., float64 (11) 
and object (14). The dataset is broken into two parts; one for 
training, which is 80\% of the dataset, and another part for 
testing, with a test ratio of 20\%. Fig.~\ref{fig: Overview of the chronic kidney disease (CKD) dataset} illustrates the 
aforementioned distribution of the dataset. 
\subsubsection{Data Preprocessing }

\begin{figure}[h!]
    \centering
    \includegraphics[width=0.8\linewidth]{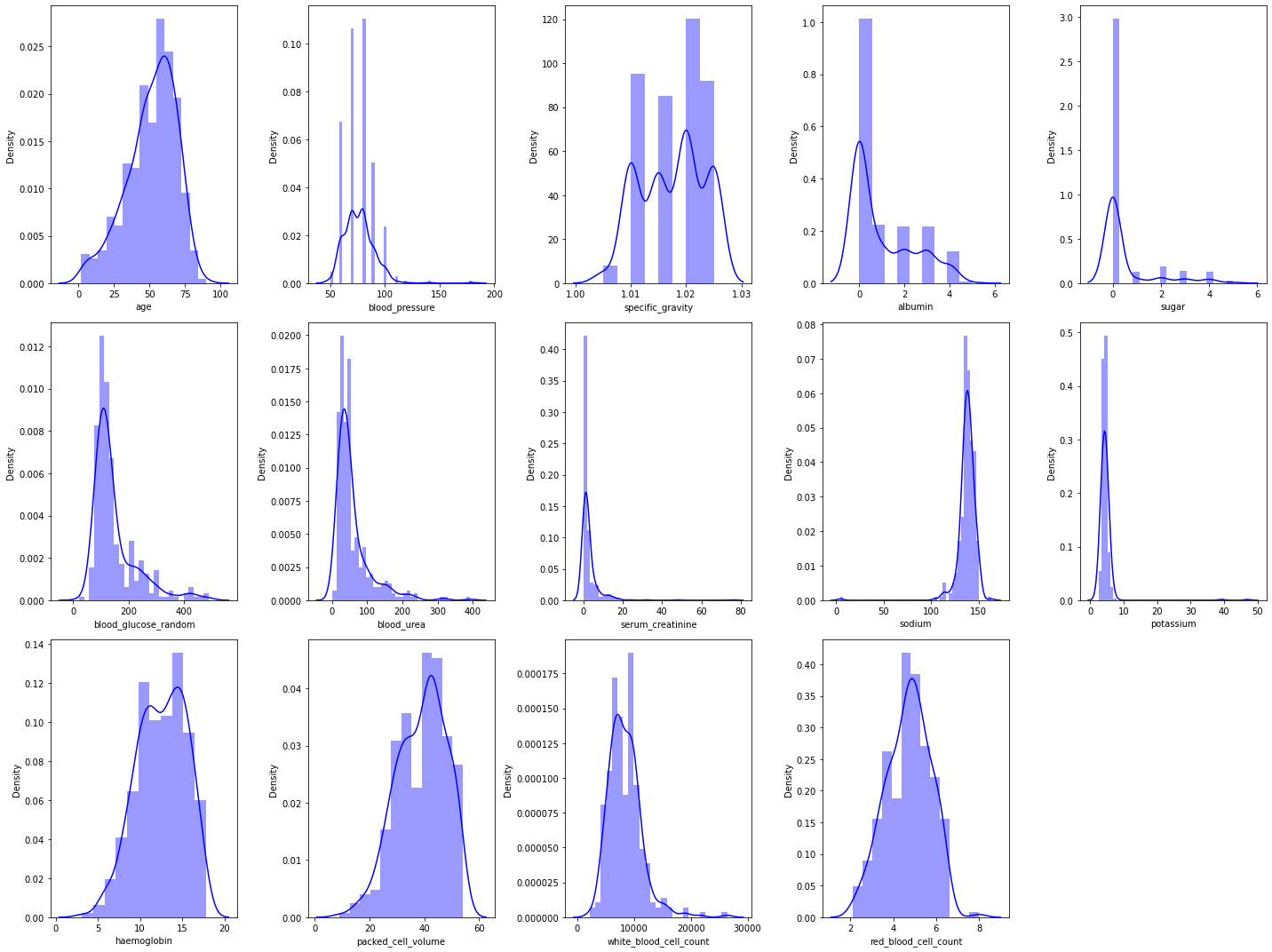} 
    \caption{ Numerical feature distribution of CKD dataset}
    \label{fig:numerical feature}
\end{figure}

It is unable to work with data that is noisy and unprocessed. 
Hence, the data needs to go through a preprocessing section, 
which contains two distinct works – ‘Formatting Data Types’ 
and ‘Feature Encoding’. As the dataset has a good number of 
missing values, however, these are fulfilled using the ‘Random 
Imputation Method’ and the ‘Mean/Mode Sampling Method’. 
In some of the columns of the dataset, we needed to converge 
the data into the corresponding datatypes. Two data frames 
have been crafted, the first of which is composed solely of 
numerical columns, and the second of which is composed 
solely of categorical columns, which is said to be the feature 
encoding. While processing, the data is renamed in order to 
make the table more user-friendly and make it simpler to deal 
with. Moreover, in the ‘classification’ column, we altered the 
‘CKD’ value to ‘0’ and the ‘not CKD’ value to ‘1’. In addition, 
we added numeric values to the entries in this column for 
future usage. Such preprocessing steps allow the machine to be 
trained more effectively. We utilized two approaches in order to 
fill in the values that were missing. Random sampling was used 
when the null value was relatively high, while the mean-mode 
sampling was used when the null value was relatively low \cite{16}. 
Following the removal of the columns containing null values, 
the dataset now contains both categorical and numerical 
columns. Machine learning cannot work with categorical data, 
thus we utilized feature encoding to transform the categorical 
columns into numeric values. The numerical feature 
distribution of the dataset is represented in Fig.~\ref{fig:numerical feature}. 

\subsection{Data Splitting }
After the completion of data processing, the dataset needed 
to be divided into two parts: train and test. To evaluate the 
performance of our models, the first thing to do is train the 
models with the training dataset and then test the models using 
the training dataset. 80\% of the dataset is considered for the 
training dataset, and the remaining 20\% will be for testing. 

\subsection{Selecting Models and Evaluation}
We choose our machine learning model and train it after 
dividing the dataset into test and training groups. The following 
models were applied: Random Forest, Support Vector Machine 
(SVM), Naive Bayes, Logistic Regression, K-Nearest 
Neighbor, XGBoost, Decision Tree, and AdaBoost. Throughout 
the evaluation phase, we evaluate each machine learning 
model's entire categorization report. In addition, the MAE and 
RMSEs were also estimated. A confusion matrix graph and 
ROC curves for each machine learning model were also 
developed in order to undertake a more detailed study of them. 
\subsection{ Machine Learning Classifiers }
\subsubsection{Decision Tree }
The Decision Tree method is a way to predict both discrete and continuous characteristics \cite{17}. It is a classification and 
regression technique. The algorithm makes predictions about 
discrete characteristics based on how the columns in a dataset 
are linked. It predicts the states of columns you tell it are 
predictable based on their values, which are called states. The 
method finds the input columns that are linked to the predicted 
column. The decision tree is easy to understand because it 
looks like the steps someone takes when making a real-life 
choice. It might help a lot when it comes to making decisions. 
It's a good idea to think about all the possible ways to solve a 
problem \cite{18}. With this method, it's not as important to clean 
the data as it is with others. 

\subsubsection{K-Nearest Neighbor} 
A straightforward supervised algorithm is the KNN \cite{19}. Both classification and regression problems can be resolved 
using it. However, classification issues are where it is most 
frequently used. KNN is a nonparametric learning algorithm 
because it does not use a specific training stage and uses all of 
the data for training, making it a lazy learning algorithm. It also 
does not take the underlying data into account. KNN stores the 
entire dataset because it lacks a model, so no learning is 
necessary. The choice of K's value is crucial because when new 
data is introduced for outcome prediction, K's neighbors are 
compared. The distance between two already labeled data is 
calculated. The distance aids in locating the new data's nearest 
neighbor. The distance is calculated using the Euclidian 
method. 

\subsubsection{Logistic Regression }
Logistic regression, a common statistical method, is used to 
model outcomes with two possible outcomes \cite{20}. Different 
learning methods are used to do logistic regression in statistical 
research. A special case of the neural network technique was 
used to make the LR algorithm. Even though this method is 
easier to set up and use, it is very similar to neural networks in 
a lot of ways. Using logistic regression, you can predict the 
outcome of a dependent variable that has a categorical value. 
So, the output must be a list of categories or a list of single 
values. There are numbers from 0 to 1 for the probabilities, but it could 
also be true or false, 0 or 1, etc. 

\subsubsection{Naïve Bayes }
The Naïve Bayes algorithm trains a classifier using the 
Bayes theorem \cite{21}. In other words, the Naïve Bayes algorithm 
was used to train a probabilistic classifier. For a specific 
observation, it determines a probability distribution across 
several classes. 

\subsubsection{Support Vector Machine }
Support Vector Machine, or SVM, is one of the most 
popular algorithms for Supervised Learning. It is used for both 
classification and regression problems \cite{22}. But in machine 
learning, it is mostly used to solve problems with classification. 
The goal of the SVM algorithm is to find the best line or 
decision boundary that divides n-dimensional space into 
classes so that we can easily put new data points in the right 
category in the future. 

\subsubsection{Random Forest }
Popular machine learning algorithms Random Forest is a 
part of the supervised learning methodology. It can be applied 
to ML problems involving both classification and regression \cite{23}. It is based on the idea of ensemble learning, which is a 
method of combining various classifiers to address complex 
issues and enhance model performance. Random Forest, as the 
name implies, is a classifier that uses several decision trees on 
different subsets of the given dataset and averages them to 
increase the dataset's predictive accuracy. Instead of relying on 
a single decision tree, the random forest uses predictions from 
each tree and predicts the result based on the votes of the 
majority of predictions. 

\subsubsection{XGBoost}
Decision trees are created sequentially in the XGBoost 
algorithm \cite{24}. Weights are significant in XGBoost. Each 
independent variable is given weight before being fed into the 
decision tree that forecasts outcomes. Variables that the tree incorrectly predicted are given more weight before being fed 
into the second decision tree. These distinct classifiers/predictors are then combined to produce a robust and accurate 
model. It can be used to solve problems involving regression, 
classification, ranking, and custom prediction. 

\subsubsection{AdaBoost}
To increase classification accuracy, boosting algorithms 
combine various weak classifiers to create strong classifiers 
\cite{25}. Adaptive Boosting, a different useful algorithm, had been 
proposed by Friedman in 1997 by Fried and Schicher, though it 
was later in 2000. It was demonstrated that LogitBoost handled 
this circumstance through improved generalizations. Numerous 
medical problems are resolved by boosting algorithms, 
including breast cancer identification, cancer detection, and 
protein structure class detection. 
\section{Results And Discussion}
During this experiment, we have been supported by the 
dataset of a famous machine learning repository of UCL. It 
includes 400 data that consists of 25 different features where 
250 data have ‘CKD’ and 150 data labeled as ‘not CKD’. The 
dataset then has been split between an 80:20 ratio for training 
and testing. After careful investigation, we have chosen several 
ML models namely SVM, Naive Bayes, Random Forest, 
Decision Tree, Ada Boost, Logistic Regression and XG Boost 
to accomplish our work.  

\subsection{Models Performance Evaluation and Comparison}
The aim of this study is to evaluate several machine 
learning techniques to classify Chronic Kidney Disease (CKD). 
Therefore, five separate evaluation metrics—Mean Absolute 
Error (MAE), F1-Score, Root Mean Square Error (RMSE), 
Precision, and Recall—are used to assess the performance of 
the models indicated above. In this section, we will discuss 
about the evaluation metrics and the result we obtained.

\begin{figure}[htbp]
    \centering
    \includegraphics[width=0.9\linewidth]{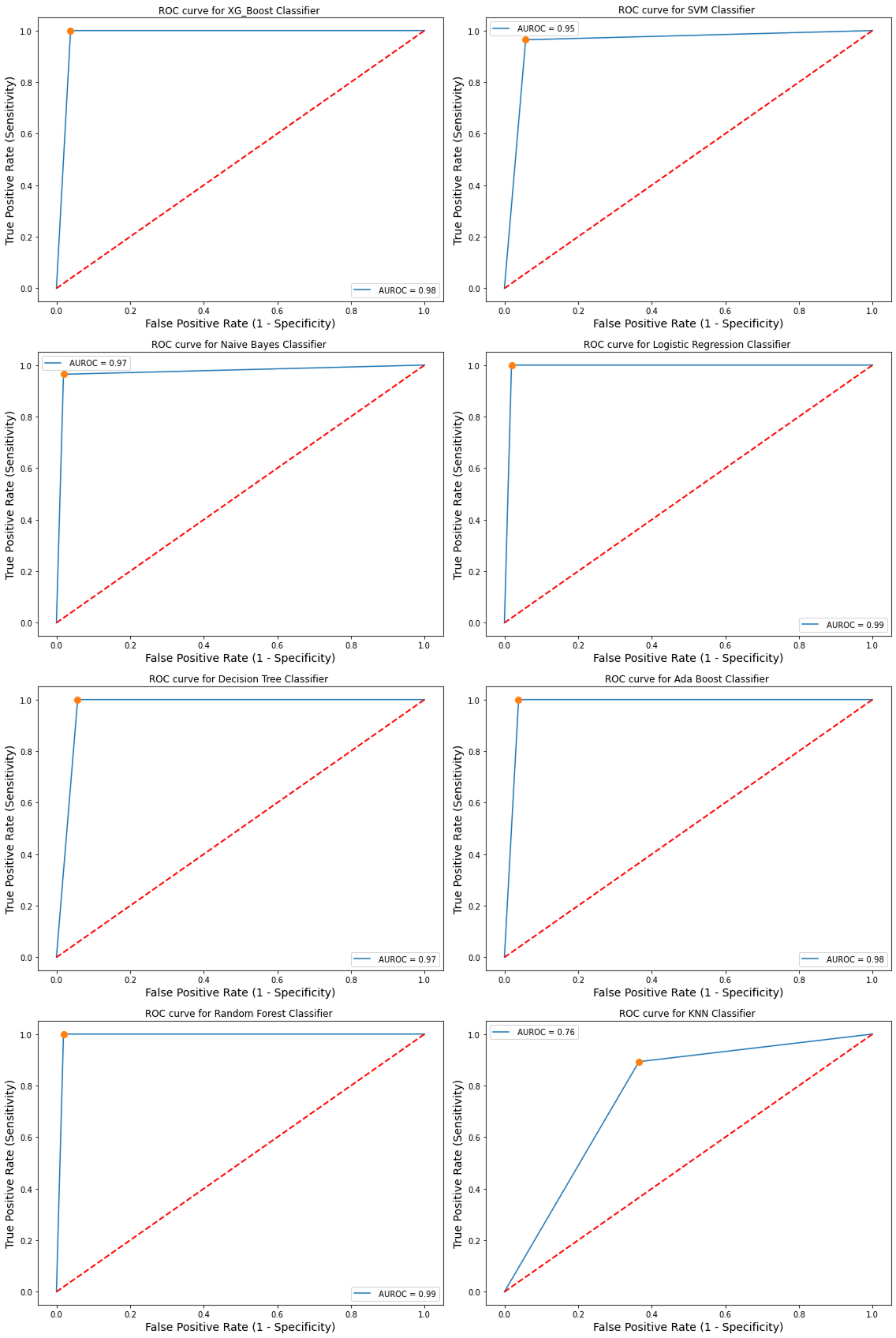} 
    \caption{ROC curve of the ML classifiers}
    \label{fig:roc_curve}
\end{figure}

In order to determine the metrics, we can identify the True 
Positive, True Negative, False Positive, and False Negative 
values by looking at the confusion matrices \cite{26} of each model, 
where positive values denote having CKD and negative values 
denote not having CKD. 

\begin{figure}[htbp]
    \centering
    \includegraphics[width=1\linewidth]{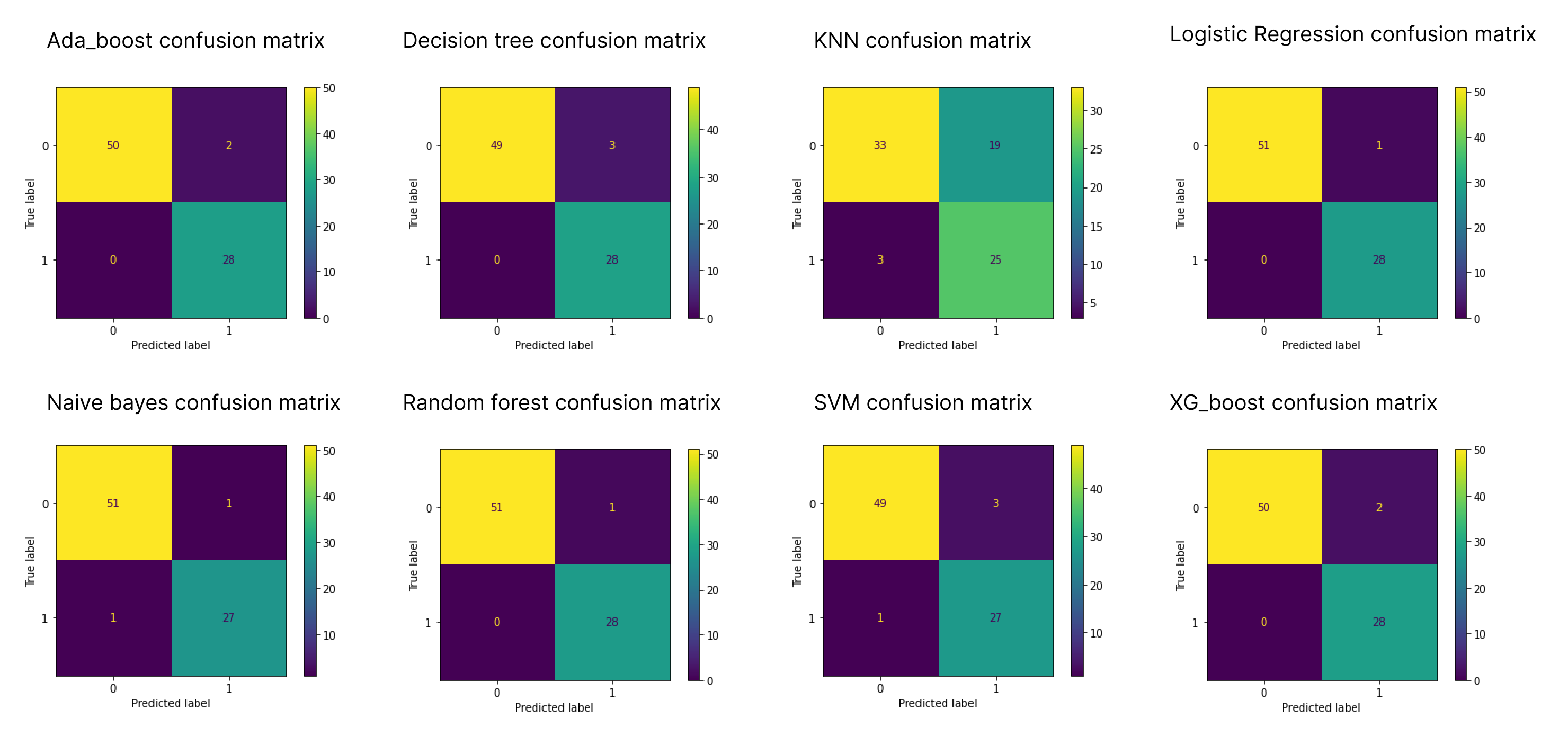} 
    \caption{ Confusion matrices of ML classifiers in CKD prediction }
    \label{fig:confusion_matrices}
\end{figure}

\begin{itemize}
    \item \textbf{TP} – Refers to a number of CKD samples that were 
accurately predicted. 
    \item \textbf{FP} – Refers to a number of projected samples that 
were thought to be CKD but aren't. 
    \item \textbf{TN} – Refers to a number of not CKD samples that 
were accurately predicted. 
    \item  \textbf{FN} – Refers to a number of samples that were CKD 
that were predicted incorrectly as not CKD. 
\end{itemize}

\begin{figure}[t]
    \centering
    \includegraphics[width=0.9\linewidth]{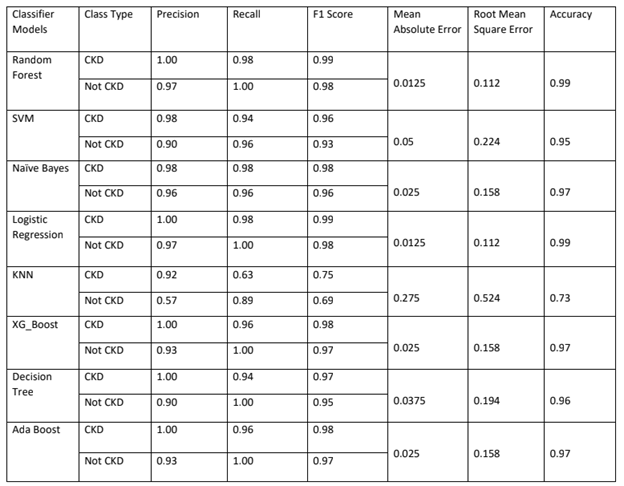} 
    \caption{Overall Performance Measure Of Ml Classifiers In Predicting CKD}
    \label{fig:overall_table}
\end{figure}

\begin{figure}[htbp]
    \centering
    \includegraphics[width=1\linewidth]{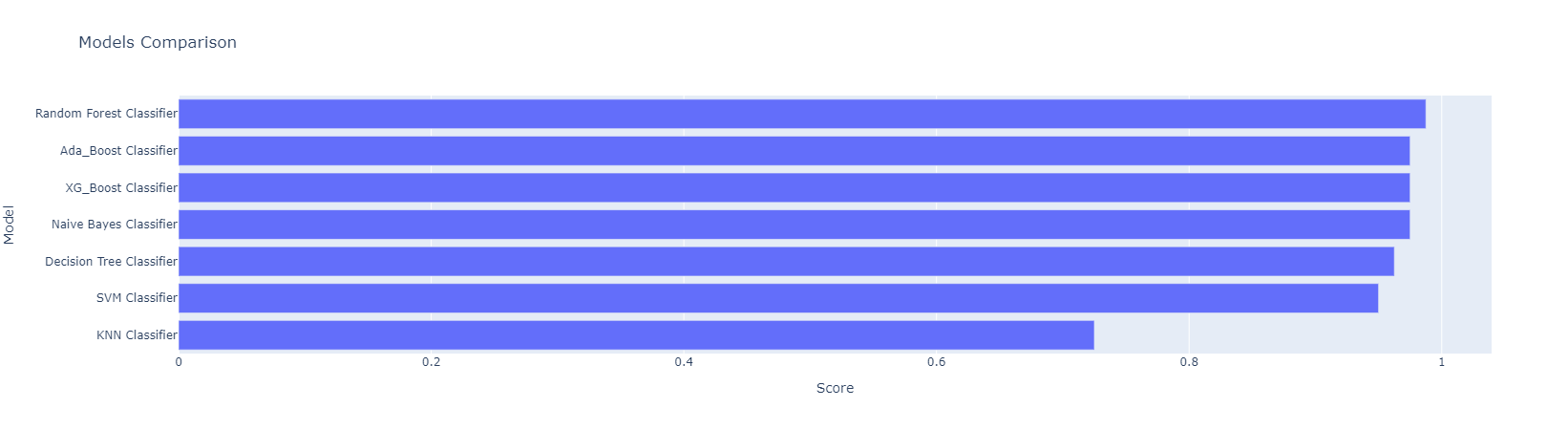} 
    \caption{Overall classification accuracies of the ML classifiers }
    \label{fig:overall_result}
\end{figure}

The proposed CKD classifiers from our experiments are determined by analyzing the ratio of prediction. Table I and Fig.~\ref{fig:overall_result} provide an overall performance insight of every model. As well as confusion matrix and ROC curve performance are shown in Fig.~\ref{fig:confusion_matrices} and Fig.~\ref{fig:roc_curve}. 

After several experiments, Table I depicts the overall performance of each supervised machine learning technique. According to the performance matrices, it is evident that most of these models except KNN performed 
relatively well on our given dataset with an accuracy well above 90\%.  

It can be identified from the performance table that Random Forest and Logistic Regression topped with the highest classification accuracy of 0.99 among other classifiers followed by the AdaBoost, XG-Boost, Naive Bayes, Decision Tree, SVM performed well with an accuracy of 0.97, 0.97, 0.97, 0.96 and 0.95 respectively. Conversely, KNN had the lowest classification accuracy which is 0.73 as well as the lowest F1- Measures compared to all other models.   

\begin{table}[htbp]
\centering
\caption{Performance Comparison with the Existing State-of-the-Art Models}
\begin{tabular}{|p{1.5cm}|p{4cm}|p{2cm}|p{4cm}|}
\hline
\textbf{Instance} & \textbf{Highest Performing Classifier(s)} & \textbf{Performance (accuracy)} & \textbf{Dataset} \\ \hline
Proposed & Random Forest, Logistic Regression & 0.99 & UCI \\ \hline
\cite{27} & Random Forest, Decision Tree & 0.95, 0.90 & Statlog heart dataset, IEEE-DataPort \\ \hline
\cite{28} & Multiclass Decision Forest, Multiclass Decision Jungle & 0.99, 0.97 & UCI \\ \hline
\cite{29} & Adam-Deep Learning, Random Forest & 0.97, 0.97 & National Kidney Foundation, Bangladesh \\ \hline
\end{tabular}
\end{table}

To identify our proposed classifiers more precisely, 
Fig.~\ref{fig:overall_result} depicts the overall performance comparison bar chart 
and Fig.~\ref{fig:confusion_matrices} depicts the confusion matrices. Observing other 
models' confusion matrices, it can be seen from Random 
Forest and Logistic Regression’s matrix that these models 
can predict TP and TN values near perfectly outperforming 
all the other models. The worst performance is shown by KNN and the matrix shows disappointing values where an 
ample amount of misclassification can be seen. 

Furthermore, the performance of the classifier can be 
widely illustrated by the ROC(Receiver Operating 
Characteristic), which is a plot of the TP rate versus the FP 
rate. So, in Fig.~\ref{fig:roc_curve}, high performing classifiers are displayed.

\subsection{Performance Comparison with the Exisitng Models }
The outcome we received from this paper has been 
compared with other models who trained in the same genre. 
From the table, we can observe a common classifier that is a 
Random forest. Its accuracy level always remains top-notch in 
different works using different datasets. Here, we just not 
compared our work with the same parameter that is CKD, 
rather we compared the outcome of our proposed work with 
different dimensions as well. Random forest performed better 
than other algorithms in a heart disease prediction study by 
Aushtmi et al. \cite{27} As a result, we may say that Random forest 
is the most effective classifier among all the classifications in 
terms of accuracy and efficiency to detect CKD.

\section{Conclusion}
In order to correctly predict CKD, our research's goal was to employ 8 different machine learning algorithms, including Random Forest, SVM, Naive Bayes, Logistic Regression, KNN, XG Boost, Decision Tree, and AdaBoost. Recall, Precision, Mean Absolute Error, Root Mean Square Error, and F1-Score were the 5 evaluation measures used to assess and compare the performance of the models. Out of all other ML techniques Random Forest 
and Logistic Regression came out as the best performer. Both these models on the given dataset provided 99\% accuracy followed by other models except KNN. KNN provided an accuracy of 73\% which is quite low compared 
to the other models. The classifier method we employed in our project has an accuracy level that meets our expectations. Before it is prepared for use in production, it still needs to undergo thorough training and testing on a variety of datasets. We hope the encouraging findings of this work will open the door to further disease prediction using ML approaches, which will subsequently help to advance healthcare systems.

\bibliographystyle{unsrt}  
\bibliography{references}

\end{document}